\documentclass{article}
\usepackage{ijcai19}

\usepackage{float}
\usepackage{threeparttable}
\usepackage{multirow}
\usepackage{times}
\usepackage{soul}
\usepackage{url}
\usepackage[hidelinks]{hyperref}
\usepackage[utf8]{inputenc}
\usepackage[small]{caption}
\usepackage{graphicx}
\usepackage{amsmath}
\usepackage{booktabs}
\usepackage{diagbox}
\usepackage{graphicx}
\usepackage{subfigure}
\usepackage{todonotes}
\usepackage{footnote}
\usepackage{amssymb}

\title{FAHT: An Adaptive Fairness-aware Decision Tree Classifier} 
\author{}
\author{
Wenbin Zhang$^1$\And
Eirini Ntoutsi$^2$
\affiliations
$^1$University of Maryland, Baltimore County, MD 21250, USA\\
$^2$Leibniz University Hannover, 30167, Hannover, Germany\\
\emails
\
wenbinzhang@umbc.edu,
ntoutsi@kbs.uni-hannover.de
}

\begin{document}

\maketitle

\begin{abstract}
Automated data-driven decision-making systems are ubiquitous across a wide spread of online as well as offline services. These systems, depend on sophisticated learning algorithms and available data, to optimize the service function for decision support assistance. However, there is a growing concern about the accountability and fairness of the employed models by the fact that often the available historic data is intrinsically discriminatory, i.e., the proportion of members sharing one or more sensitive attributes is higher than the proportion in the population as a whole when receiving positive classification, which leads to a lack of fairness in decision support system. A number of fairness-aware learning methods have been proposed to handle this concern. However, these methods tackle fairness as a static problem and do not take the evolution of the underlying stream population into consideration. In this paper, we introduce a learning mechanism to design a fair classifier for online stream based decision-making. Our learning model, \textsf{FAHT}
(Fairness-Aware Hoeffding Tree), is an extension of the well-known Hoeffding Tree algorithm for decision tree induction over streams, that also accounts for fairness. 
Our experiments show that our algorithm is able to deal with discrimination in streaming environments, while maintaining a moderate predictive performance over the stream.
\end{abstract}

\section{Introduction}

Information systems are increasingly becoming automated and data-driven in both online and offline settings to render all sorts of decisions, including the allocation of resources, issuing of mortgage loans, assessment of pretrial risk, personalizing marketing and pre-screening and acceptance of applicants. However, as the usage of automated decision-making systems in assistance or even replacement of human-based decision making is widely adopted, growing concerns have been voiced from civil organizations, policy-makers and legal experts about the potential loss of fairness and accountability in the employed models~\cite{datta2015automated,sweeney2013discrimination,zliobaite2015survey}.

Data-driven algorithms build simple or complex hypotheses based on historical data in order to make future decisions. However, even if the algorithm itself is non-discriminatory, the trained models can still be discriminative due to the naturally inherited biases encoded in the historical data. Simply making the decision-making algorithm blind to the sensitive attribute(s) does not work as other attributes may be correlated with the suppressed attribute(s). Such indirect and discriminative relations inherited from the historical data can therefore still persist in future predictions. The basic aim of fairness-aware classifiers is to make fair and accurate decisions, i.e., to train a decision-making model based on biased historical data such that it provides accurate predictions for future decision-making, yet does not discriminate deprived subgroups in the population.

This increasing attentive issue has motivated a number of approaches, ranging from discrimination discovery and elimination to discrimination interpretation, to ameliorate the intrinsic bias of the historical data in order to provide fairness-aware decision-making systems. However, all of these studies tackle fairness as a static problem, assuming that the characteristics of the underlying population do not evolve. In many real-world applications though data is generated in a streaming fashion and its characteristics might vary over time. Therefore, discrimination-aware learning for such sort of applications should be able to cope with changes by adapting the model to the underlying distribution. 

In order to simultaneously address these challenges, we introduce a new fairness-aware classifier for data streams. The main contributions of this paper are: i) We reformulate information gain for the fairness-aware classification setting. Specifically, the fair information gain splitting criterion is introduced  that jointly considers the information and the fairness gain of a split.
ii) An adaptive fairness-aware learner is designed for online stream classification, which is able to achieve a good predictive performance over the stream while also preserving a low discrimination score.
iii) A detailed experimental evaluation is provided that shows the superior predictive performance vs fairness behavior of our model. Moreover, the impact of the proposed fairness-aware splitting criterion on the structural properties of the resulting model are also investigated. \emph{To the best of our knowledge, this is the first work to address discrimination in data stream classification}.

The remainder of the paper is organized as follows. Section~\ref{sec: background} and~\ref{sec: relatedWork} review the theoretical background knowledge regarding discrimination-aware learning and relevant work. Section~\ref{sec: method} proposes the learning mechanism for fairness-aware classification in data streams. The experimental results are presented and analyzed in Section~\ref{sec: experiment}. Finally, we conclude the paper in Section~\ref{sec: conclusion}.

\section{Basic Notions}
\label{sec: background}
Assume a set of attributes \{$A_1,\cdots,A_d$\} with their respective domains $dom(A_i)$ and let $Class$ be the class attribute. The data stream $S$ consists of instances over the schema $(A_1,\cdots,A_d, Class)$ arriving over time, 
with each instance $x\in S$ being an element of $dom(A_1)\times \cdots dom(A_d)$ and class label $c \in dom(Class)$.
Our learning setting is fully-supervised, however, the label of an instance $x$ arriving at time $t$ becomes available shortly after its arrival. Therefore, the goal is to make a prediction for $x$ at $t$ using
the current classifier and then, upon the arrival of its label to use
the labeled instance $(x, c), c \in dom(Class)$, for model update. This setup is known
as first-test-then-train or prequential evaluation.

We assume a special attribute $SA$, referred as $sensitive~attribute$ with a special value $s \in dom(SA)$ referred as $sensitive~value$ that defines the discriminated community. 
For simplicity, we assume that $SA$ is a binary attribute: $dom(SA)=\{s, \overline s \}$. As a running example, we use $SA=$``gender'' as the sensitive attribute and $s=$``female'' as the sensitive value (with $\overline s=$``male''). We also assume the class is binary with values \{$rejected, granted$\}.
By combing $SA$ and $Class$ attributes (both binary), four communities are created:
\begin{itemize}
	\item \textbf{DR} (deprived-rejected): females rejected a benefit.
	\item \textbf{DG} (deprived-granted): females granted a benefit.
	\item \textbf{FR} (favored-rejected): males rejected a benefit.
	\item \textbf{FG} (favored-granted): males granted a benefit.
\end{itemize}
A large number of fairness-related measures has been proposed in the literature, see \cite{verma2018fairness}. In this work, we adopt the widely used \emph{statistical parity}  that examines whether the probability of being granted is the same for both deprived and favored communities. More formally, for a set of instances $D$ from the given problem:
\begin{equation}
\label{equ: discrimination}
Disc(D)= \frac{FG}{FG+FR}- \frac{DG}{DG+DR}
\end{equation}
If more of people of the deprived community are rejected a benefit comparing to the people of the favored community, the former could claim that they are discriminated.
Statistical parity can be evaluated in the original dataset $D$ or in the predictions of a classifier, capturing the fairness of the dataset representation or of the classifier, respectively.

\section{Related Work}
\label{sec: relatedWork}
Relevant to our work is work on fairness-aware learning as well as works on stream learning, in particular stream classification with decision tree models.

\subsection{Fairness-aware Learning}
A number of research approaches have been proposed to address the problem of bias and discrimination in machine learning systems due to the inherent bias in data and the complex interaction between data and learning algorithms~\cite{pedreshi2008discrimination,zliobaite2015survey,hajian2016algorithmic}. 
These approaches generally fall into one of the following categories: i) pre-processing approaches, ii) in-processing approaches and iii) post-processing approaches.

\emph{Pre-processing approaches} modify the data distribution to ensure a fair representation of the different communities (i.e., DR, DG, FR, FG) in the training set. The rationale for these approaches is that if a classifier is trained on discrimination-free data, its predictions will not be discriminatory (such a property however, cannot be guaranteed). 
A popular method in this category is massaging~\cite{kamiran2009classifying} that swaps the class labels of selected instances to restore balance. A ranker is used to carefully select the instances to be swapped in order to minimize the effect of label swapping on predictive accuracy; these are the instances closest to the decision boundary. 
Instead of intrusively relabeling the instances, reweighting~\cite{calders2009building} assigns different weights to different communities to alleviate the bias towards favored communities for the sake of benefiting the classification of deprived communities. The weights are assigned based on the expected and observed probability of an instance according to its sensitive attribute value and class. A higher weight will be assigned if the observed probability is lower than the expected probability to neutralize the discrimination. 
We emphasize that pre-processing approaches are not effective in eliminating discrimination arising from the learning algorithm itself. 

\emph{In-processing approaches} modify existing learning algorithms to also account for fairness instead of only predictive performance. Therefore, distinct from the first strategy which is classifier-agnostic, these approaches are algorithm-specific. 
For example, \cite{aghaei2019learning} added regularization terms to the Mixed-Integer Programming model to penalize discrimination. \cite{calders2010three} proposes three approaches for fairness-aware Naïve Bayes classifiers: the first one alters the decision distribution until there is no more discrimination, the second one attempts to remove the correlation between sensitive attribute and class label by building a separate model for each sensitive group, and the third one models a latent variable to discover the actual class labels of a discrimination-free dataset. 
Closer to our work is the work  by~\cite{kamiran2010discrimination} that incorporates discrimination into the splitting criterion of a decision tree classifier. Two key distinctions are: i) fairness in our work is directly defined in terms of the discrimination difference of the induction of a split, i.e., the fairness gain due to the split rather than the entropy w.r.t. sensitive attribute and ii) our model operates in an online setting rather than upon a static/batch dataset. The online learning scenario is much more challenging as decisions have to be made without access to the complete dataset and moreover, such decisions have an impact on following decisions in terms of both accuracy and fairness (feedback loops).

Finally, \emph{post-processing approaches} modify the resulting models by ``correcting'' the decision regions that lead to redlining for a fair representation of different subgroups in the final decision process. In~\cite{hajian2015discrimination}, the fair patterns are processed with $k$-anonymity to work against discrimination.  \cite{kamiran2010discrimination} carefully relabels selected leaves of a decision tree model to reduce  discrimination with the least possible impact on accuracy.
Transferring such approaches to a stream setting is not straightforward as the decision regions/leaves might change over the course of the stream due to concept drifts.

\subsection{Stream Classification}
The main challenge for learning in a stream environment is the so called, concept-drifts, i.e., changes in the joint data distribution over time~\cite{krawczyk2017ensemble,aggarwal2007data,gama2014survey}. The learning methods therefore should be able to adapt to changes by learning incrementally from new instances, e.g.,~\cite{oza2005online,zhang2017hybrid} and by carefully considering historical information into the model by e.g., forgetting outdated information, e.g.,~\cite{melidis2018learning,wagner2015ageing} and/or focusing on most recent data, e.g.,~\cite{bifet2007learning}


Our approach integrates the algorithm-level solution for fairness-aware classification and the online approach to maintain an accurate and up-to-date classifier for infinite data streams with non-stationary distribution and discrimination bias. We build our model on top of the Hoeffding tree~\cite{domingos2000mining}, one of the most popular decision tree classifiers for data streams. Hoeffding tree relies on the Hoeffding bound to make confident decisions about the selection of splitting attributes over infinitive streams and accommodates new instances from the stream incrementally. 
To enable fairness-aware learning, we revise the learning model by not only incorporating new examples from the stream but also by alleviating their bias towards the favoured group through a new splitting criterion, the \emph{fair information gain}, that jointly considers the information and the fairness gain of a split.

\section{Fairness-Aware Hoeffding Tree Classifier}
\label{sec: method}
This section first discusses vanilla Hoeffding tree, then the fair information gain splitting criterion for the discrimination-aware tree induction is introduced followed by the specification of a number of refinements and modifications that instantiate the fairness-aware learning.

\subsection{Vanilla Hoeffding Tree (HT)}

Our Fairness-Aware Hoeffding Tree (FAHT)  classifier builds upon the Hoeffding tree classifier~\cite{domingos2000mining}. As discussed in Section \ref{sec: relatedWork}, learning from data streams requires a learning algorithm that is able to adapt to the changes in the underlying data distribution. The Hoeffding tree (HT)~\cite{domingos2000mining} achieves that by making confident attribute splitting decisions in an online fashion. Given a stream of instances, the HT induction algorithm inspects each instance in the stream only once and stores sufficient information in the leaves of  the  tree, that allow tree growing. The crucial decisions needed for tree growing are when and how to split a node. To this end, the authors employ the Hoeffding bound~\cite{maron1994hoeffding} to guarantee that a splitting decision is asymptotically nearly identical to the decision of a conventional static learner. Such decisions aim to optimize for predictive performance and are based on information gain; as such, original HT does not consider fairness.

In this work, we extend the HT model in two ways: i) by introducing a new splitting criterion that \emph{jointly} considers the gain of an attribute split w.r.t. classification and also w.r.t. discrimination (Section~\ref{sec:FAHT-splitting}) and ii) by maintaining sufficient statistics at each node to enable the computation of the new splitting criterion values (Section~\ref{sec:FAHT-statistics}).

\subsection{The Fair Information Gain Splitting Criterion}
\label{sec:FAHT-splitting}
The HT attribute-splitting decisions are exclusively accuracy-oriented and therefore an unfair tree might be induced over the course of the stream. To overcome this, we propose to alter the splitting criterion to also consider the fairness gain of a potential split.
To this end, we first define the \emph{fairness gain} of an attribute $A$ relative to a collection of instances $D$ as the discrimination reduction in $D$ due to splitting on $A$:
\begin{equation}
\label{equ: DL}
FG(D,A)= |Disc(D)|-\sum_{v \in dom(A)}\frac{|D_v|}{|D|}|Disc(D_v)|
\end{equation}


\noindent where $D_v, v \in dom(A)$ are the partitions induced by $A$. 
In practice, the discrimination before splitting (i.e., on $D$) is compared against the discrimination after splitting, by aggregating the discrimination on each resulting subset $D_v$. The corresponding discrimination values for each partition are computed based on Equation~\ref{equ: discrimination}.

The idea of fairness gain (FG) aligns with the idea of information gain (IG) which measures the reduction in entropy resulting from a split; for both FG and IG it holds that the higher the reduction the merrier. The distinction is that the merit of $FG$ is evaluated from the discrimination perspective while $IG$ pays attention to accuracy. In addition, different from the previous work, $FG$ is directly defined in terms of the reduction in discrimination due to a split rather than mediating between the entropy w.r.t. sensitive attribute as in~\cite{kamiran2010discrimination}. The limitation of such a mediation was acknowledged in their paper; in particular, their discrimination-aware splitting criteria does not lead to significant discrimination reduction unless being used in conjunction with additional leaf relabeling operations, thus combining in-with post-processing fairness interventions.

We combine fairness gain and information gain to a joint objective, called \emph{fair information gain} (FIG), that evaluates the suitability of a candidate splitting attribute in terms of both predictive performance and fairness. More formally, the fair information gain of an attribute $A$ relative to a set of instances $D$ is defined as follows:

\begin{equation}
\footnotesize
\label{equ: ConjFairInfo}
FIG(D,A) = \left\{
	\begin{array}{lr}
IG(D,A) & ,if FG(D,A)=0 \\
IG(D,A) \times FG(D,A) & ,otherwise
	\end{array}
\right.
\end{equation}

Intuitively, the fair information gain closely combines information gain $IG$ and fairness gain $FG$. For attributes that their split does not change the distribution of discrimination, i.e., $FG=0$, their $FIG$ is reduced to $IG$. The iterative tree induction process therefore takes into account the influence of the splits under evaluation on the class distribution as well as the discrimination of the resulting tree, i.e., the construction of the tree is both accuracy- and fairness-driven. 

Note that the trade-off between accuracy and discrimination can be also achieved by combining $IG$ and $FG$ through other operations, e.g., addition. However, these two metrics are not necessary in the same scale in practice and the values of them can be incomparable, which means one can be dominated by the other. Multiplication is therefore favoured over addition. Another merit of multiplication is that it encourages fair splits by giving priority to splitting candidates that result in the reduction of discrimination over those that increase discrimination, i.e., when $FG$ becomes negative. This decision is also verified experimentally.

\subsection{FAHT: A Fairness-aware HT Classifier}
\label{sec:FAHT-statistics}
Our fairness-aware decision tree induction algorithm for streams is built on top of the vanilla HT using the newly introduced fair information gain splitting criterion (c.f., Section~\ref{sec:FAHT-splitting}) that aims at optimizing both predictive performance and fairness of splits.
Moreover, several refinements and modifications to the original HT are required to allow for a fairness-aware Hoeffding tree learner over streams. These extensions are described hereafter. Note that other designs and theoretical guarantees still hold up and theorems can be proven accordingly. \\

\noindent
\textbf{Pre-pruning.} In the original HT algorithm, the idea of it might be more beneficial to not split a node at all is carried out by also considering the merit of no split, represented by the null attribute $X_\emptyset$ at each node for pre-pruning. Thus, a node is only allowed to split when the best split found is sufficiently better, according to the same Hoeffding bound test that determines differences between other attributes, than $X_\emptyset$. In our implementation, the merit to be maximized is the fair information gain. Therefore, the $FIG$ of the best candidate split should be sufficiently better than that of the the non-splitting option. In terms of the $FIG$ of the null attribute, the current class distribution is used to represent the $IG$ and the $FG$ is evaluated as the current level of discrimination.\\

\noindent
\textbf{Sufficient statistics.} The original HT algorithm scans each instance in the stream only once and stores sufficient information in its leaves to enable the calculation of the information gain for each possible split. In FAHT, additional information that necessitates the calculation of fair information gain should be also stored. 
In case of \emph{discrete attributes}, each node in the tree maintains a separate table per attribute. The counts of the class labels that apply for each attribute value are stored to calculate the information gain. In addition, the counts of each attribute value that keep track of the numbers of deprived and favored instances as well as of the instances receiving positive classification in deprived and favored groups are maintained to evaluate the fairness gain afforded by each possible split. The appropriate entries are updated incrementally based on new instances from the stream, based on attribute values, sensitive attribute values and class values. In case of \emph{numeric attributes}, each class label maintains its sufficient statistics as a separate Gaussian distribution. So are the four previous mentioned discrimination calculation related statistics. The updating of the numeric attribute involves updating appropriate distribution statistics according to the sensitive attribute values and class of the continuously arriving examples. The allowing test is used to select potential thresholds for binary splits and the merit of each allowed threshold is evaluated to select the most appropriate split among them. The merit is also computed according to the proposed fair information gain.\\

\noindent
\textbf{Memory.} 
The counts stored in the leaves are the sufficient statistics needed to compute the information gain afforded by each possible split. However, efficient storage is important.
If there are $d$ attributes with a maximum number of $v=max_{A_i}|dom(A_i)|, 1 \leq i\leq d$ values per attribute and $c=|dom(Class)|$ possible classes in total, then a memory of $O(dvc)$ is required to store the necessary sufficient statistics at each leaf of the original HT algorithm. 
Our FAHT requires the maintenance of the four previous mentioned extra quantities in order to compute the proposed fair information gain measure. This is basically equivalent to add two more attributes, each with two values. Thus, the required memory becomes $O((d+2)vc)$ and therefore FAHT incurs negligible extra costs especially when $d \gg 2$. In the like manner, the continuous numeric attributes can also be stored efficiently for a slightly additional memory to compute the proposed heuristic measure afforded by each possible binary split.

\section{Experiments}
\label{sec: experiment}
The first goal of our experiments is to evaluate 
the predictive- vs fairness-performance of our proposed FAHT method, as is typical in the domain of fairness-aware machine learning~\cite{verma2018fairness}. 
To this end, we evaluate the different models in terms of accuracy and statistical parity.
Due to the streaming nature of the data we use prequential evaluation, that is for each incoming instance from the stream we first predict its class via the model before updating the model ~\cite{gama2010knowledge}. We report on both aggregated measures as well as on the over the stream performance of the different methods ~(Section~\ref{sec:exp_accu_fair}). Except for the landmark window model employed by the original HT, we also report on a sliding window model variation that focuses on the most recent history from the stream.

A second goal of our experiments is to understand the effects of our proposed fair information gain splitting criterion in the structure of the resulting decision tree models.
To this end, we assess the impact of the new proposed FIG measure into the selected splitting attributes and eventually, in the tree structure as well as the latency in the splitting attribute selection process (Section~\ref{sec:exp_structure}). 


%


\subsection{Datasets}
Despite the growing concern on the discriminative behavior of AI models, there is still a lack of datasets and benchmarks that will allow for the development of new fairness-aware methods as well as for a systematic assessment of their capabilities~\cite{hajian2016algorithmic}. With respect to fairness-aware learning in data streams, this challenge is further magnified by the demanding requirement for big non-stationary datasets. 
Among the available datasets, the ones that best meet our requirements are the \emph{Adult}  and the \emph{Census} datasets~\cite{Dua:2017} both referring to the same learning task of predicting whether the annual income of a person will exceed 50K dollars.

The \emph{Adult} dataset consists of 48,843 instances, each instance modeling a person in terms of 14 employment and demographic attributes. The sensitive feature is ``gender'' with female being the deprived group and male being the favored group. Making an annual income of more than 50K dollars is considered as receiving positive classification. The intrinsic discrimination level of the dataset is 19.45\%, based on Equation~\ref{equ: discrimination}.
The \emph{Census} dataset has the same learning task but is considerably larger with 299,285 instances and 41 attributes. The discrimination in the whole dataset is 7.63\% according to Equation~\ref{equ: discrimination}. Existing works address these datasets from a static learning perspective. We render them as discriminated data streams by randomizing the order of the instances and processing them in sequence.

\subsection{Accuracy vs. Fairness}
\label{sec:exp_accu_fair}
To the best of our knowledge, this is the first work to address discrimination in data stream classification, so we compare our both accuracy- and fairness-driven FAHT to the pure accuracy-driven HT. As a second baseline, we incorporate into the HT model the static discrimination-aware splitting criterion of \cite{kamiran2010discrimination}, we refer to it as ``Kamiran's'' method. The obtained results are summarized in Table~\ref{table:InstanceBasedperformance}.

\begin{table}[!htb]
	\centering
	\setlength{\tabcolsep}{1pt}
	\renewcommand\arraystretch{}
	\resizebox{0.48\textwidth}{7mm}{
		\begin{tabular}{|c|c|c|c|c|}
			\hline
			\multirow{2}{*}{\diagbox{Methods}{Metric}}& \multicolumn{2}{c|}{\textbf{Adult dataset}} & \multicolumn{2}{c|}{\textbf{Census dataset}} \\ 
			\cline{2-5}
			& Accuracy &  Discrimination  & Accuracy & Discrimination  \\
			\hline
			\textbf{HT} & 83.91\%  &  22.59\%  & 95.06\% & 6.84\%  \\
			\hline
			\textbf{Kamiran's} & 83.92\% (\textbf{+0.01\%})  & 22.61\% (\textbf{+0.09\%}) & 94.82\% (\textbf{-0.25\%}) & 6.59\% (\textbf{-3.65\%}) \\
			\hline
			\textbf{FAHT} & 81.83\% (\textbf{-2.48\%})  &16.29\% (\textbf{-27.89\%}) & 94.28\% (\textbf{-8.20\%})  & 3.20\% (\textbf{-53.22\%}) \\
			\hline
	\end{tabular}}
	\caption{Accuracy-vs-discrimination evaluation of different models.}
	\label{table:InstanceBasedperformance}
\end{table}

In these results, it is clear that our model is capable of diminishing the discrimination to a lower level while maintaining a fairly comparable accuracy. In particular, FAHT achieves a discrimination decrease of 27.89\% and 53.22\% at the cost of a slight 2.48\% and 8.20\% accuracy reduction on Adult and Census dataset, respectively. Due to the exclusively accuracy-oriented tree construction and the intrinsic discrimination bias of the historic data, a lack of fairness tree can be induced during the construction of the HT. Regarding Kamiran's method, little numerical difference is observed comparing to HT. The resulting trees of the methods look also pretty similar\footnote{Ommitted due to lack of space.} and therefore, we do not include it in the following discussion.
It is however clear that our method that seamlessly integrates the fairness merit into the tree induction, results into a model that is both accuracy-and fairness-driven. This anti-discrimination capability of FAHT is also statistically significant as shown in Table~\ref{table:McNemarTest}. 


\begin{table}
	\centering
	\begin{threeparttable}[!htb]
		\label{tab:test2}
		\resizebox{0.47\textwidth}{7mm}{
		\begin{tabular}{|c|c|c|c|c|}
			\hline
			\multirow{2}{*}{\diagbox{\textbf{HT}}{\textbf{FAHT}}}& \multicolumn{2}{c|}{{\textbf{Adult dataset}}$^1$} & \multicolumn{2}{c|}{{\textbf{Census dataset}}$^2$}\\ 
			\cline{2-5}
			& Granted &  Rejected & Granted & Rejected\\
			\hline
			Granted &  527 & 310 & 824 & 963 \\
			\hline
			Rejected &   523 & 14,832 & 564 & 153,424 \\
			\hline
		\end{tabular}}
		\begin{tablenotes} 
			\item[1] Chi-squared = 53.954, df = 1, p-value = 2.052e-13 
			\item[2] Chi-squared = 103.74, df = 1, p-value $\textless$ 2.2e-16
		\end{tablenotes}
	\end{threeparttable}
	\caption{McNemar's test on deprived community between HT and FAHT applied to each dataset, testing whether FIG worked to benefit the positive classification of the deprived group.}
	\label{table:McNemarTest}
\end{table}

\begin{figure}[!t]
	\centering
	\vspace{-2mm}
	\makebox[0.6\textwidth][l]{
		\subfigure[Accuracy]{
			\includegraphics[height=0.2\textheight,width=0.25\textwidth]{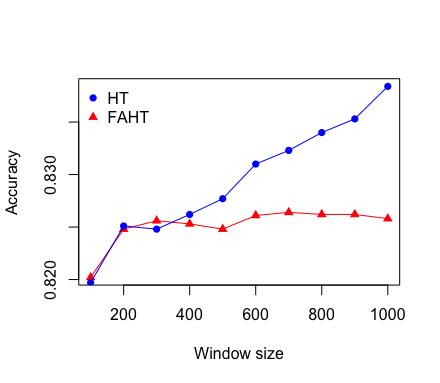}
		}%
		\subfigure[Discrimination]{
			\includegraphics[height=0.2\textheight,width=0.25\textwidth]{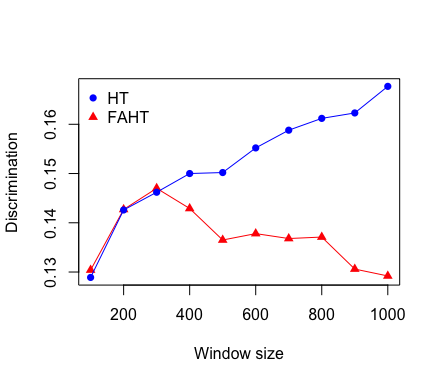}
		}%
	}
	\centering
	\caption{The comparison of performance metrics in different data stream strategies. The Adult data stream is processed in sliding windows; Each window trains a base learner as the ensemble component, the oldest one will be replaced when the classifier window is full; The ensemble members stored in the classifier window will also get updated with the instances in the current sliding window.}
	\label{fig:windowEnsembleResults}
\end{figure}

FAHT learns incrementally over the course of the stream. To evaluate the applicability of the proposed fairness enabled learning and how discrimination is propagated due to the stream processing, we employ FAHT as a base learner in the other two data stream adaption categories (we will now focus on the Adult dataset as the illustrating example due to space constraints\footnote{The entire experimental results and code are available at \\ ~~~~~~~~~~~~~~\url{https://github.com/vanbanTruong/FAHT}}). Specifically, the data stream is processed using sliding windows and a queue structured classifier window is maintained as the collection of learners from the previous streams. A new learner is trained based on the most recent sliding window and will be added to contribute for the ensemble learning. Again, the HT and FAHT are employed as the based learners for comparison. During the learning process, the components of the ensemble classifier are trained incrementally with the instances in the new sliding window. The obtained results are depicted  in Figure~\ref*{fig:windowEnsembleResults}. 

As one can see, FAHT consistently pushes the discrimination to a lower level while maintaining a fairly comparable accuracy when being used as the base learner in the other two data stream adaption categories. In particular, FAHT achieves the best discrimination reduction while HT gives the worst fairness result  when the window size is 1000. This is not surprising; with the updates discrimination inherited from the previous stream could propagate and persist in later predictions. Therefore, although HT provides a better prediction performance, it leads to an unfair  model. On the other hand, the fair information gain splitting criterion of FAHT helps to control the discrimination propagation and manages to  push the discrimination to a low level while maintaining a high prediction capability. These results are consistent with the results from the previous incremental-learning fashion and show the discrimination-aware learning ability of the proposed method.
Therefore, the applicability and anti-discrimination propagation merit of our fairness-aware data stream classifier hold.

\subsection{Structural Effects on the Tree Construction}
\label{sec:exp_structure}
In this section, we further investigate the accuracy-driven and fairness-oriented construction of FAHT. Table~\ref{table:attCorr} shows the Pearson correlation coefficients between each attribute as well as class label of the experimental dataset. Note that, for space constraint, only attributes being used as splitting attributes during the constructions of both HT and FAHT methods are shown, and correlation between each pair of variables is computed using all complete pairs of observations on those variables. In addition, different from static decision tree induction, an attribute will not be relaxed after being selected as the splitting attribute at one particular node and is a candidate splitting attribute for the succeeding splitting selection as well. In the constructed HT, ``capital-gain'' is selected as the root node and is re-selected several times at the succeeding nodes. This is quite intuitive as capital-gain is directly related with the amount of annual salary, i.e., class label. However, ``capital-gain'' also mirrors the intrinsic discrimination bias of the historic data as male, the favored community, has a higher likelihood of receiving higher capital-gain than female, the deprived community, based on the training data.
On the other hand, FAHT selects ``age'' as the root node and re-selects it several times at the succeeding nodes as well. Generally speaking, income per annum increases as one gets older and more experienced, which holds regardless of the sensitive attribute value and from the correlation analysis. Selecting such an attribute balances encoding and diminishing discrimination of the training data for an accuracy-driven as well as fairness-oriented model. Such attribute selection strategy can also be concluded from Table~\ref{table: boundaryCorr}, which shows the Pearson correlation coefficients between sensitive attributes and decision boundaries. As one can see, the sensitive attribute is more correlated with the predicted boundary of HT than FAHT's, because our model tries to build the boundary that the proportion of each community when receiving positive classification is identical or differs slightly, i.e., non-discriminatory. We also observe that the correlation between predicted boundary and actual boundary is stronger in HT than FAHT. This is because FAHT manages the trade-off between accuracy and discrimination when building the boundary while HT is completely accuracy-driven.

\begin{table}[!htb]
	\centering
	\setlength{\tabcolsep}{1pt}
	\renewcommand\arraystretch{}
	\resizebox{0.48\textwidth}{20mm}{
		\begin{tabular}{|c|c|c|c|c|c|c|c|c|c|c|c|}
			\hline
			\tiny{\textbf{Attribute}} & \tiny{age}& \tiny{education}& \tiny{marital.status} & \tiny{relationship}& \tiny{sex} & \tiny{capital.gain} & \tiny{capital.loss}& \tiny{hours.per.week}& \tiny{native.country}& \tiny{class}\\
			\cline{1-11}
			\tiny{age} & \tiny{1.00} & \tiny{0.01} & \tiny{0.43}& \tiny{-0.22} &\tiny{-0.09}& \tiny{0.08}& \tiny{0.06}& \tiny{0.07}& \tiny{-0.01}& \tiny{0.23}\\
			\cline{1-11}
			\tiny{education} & \tiny{0.01} & \tiny{1.00} & \tiny{0.01}& \tiny{0.05} &\tiny{0.00}& \tiny{0.02}& \tiny{-0.01}& \tiny{-0.05}& \tiny{0.06}& \tiny{-0.05}\\
			\cline{1-11}
			\tiny{marital.status} & \tiny{0.43} & \tiny{0.01} & \tiny{1.00}& \tiny{0.02} &\tiny{0.18}& \tiny{0.01}& \tiny{0.01}& \tiny{0.01}& \tiny{0.00}& \tiny{0.00}\\
			\cline{1-11}
			\tiny{relationship} & \tiny{-0.22} & \tiny{0.05} & \tiny{0.02}& \tiny{1.00} &\tiny{0.27}& \tiny{-0.04}& \tiny{-0.05}& \tiny{-0.18}& \tiny{0.04}& \tiny{-0.17}\\
			\cline{1-11}
			\tiny{sex} & \tiny{-0.09} & \tiny{0.00} & \tiny{0.18}& \tiny{0.27} &\tiny{1.00}& \tiny{-0.05}& \tiny{-0.05}& \tiny{-0.23}& \tiny{0.00}& \tiny{-0.21}\\
			\cline{1-11}
			\tiny{capital.gain} & \tiny{0.08} & \tiny{0.02} & \tiny{0.01}& \tiny{-0.04} &\tiny{-0.05}& \tiny{1.00}& \tiny{-0.03}& \tiny{0.08}& \tiny{-0.01}& \tiny{0.22}\\
			\cline{1-11}
			\tiny{capital.loss} & \tiny{0.06} & \tiny{-0.01} & \tiny{0.01}& \tiny{-0.05} &\tiny{-0.05}& \tiny{-0.03}& \tiny{1.00}& \tiny{0.05}& \tiny{0.00}& \tiny{0.15}\\
			\cline{1-11}
			\tiny{hours.per.week} & \tiny{0.07} & \tiny{-0.05} & \tiny{0.01}& \tiny{-0.18} &\tiny{-0.23}& \tiny{0.08}& \tiny{0.05}& \tiny{1.00}& \tiny{-0.01}& \tiny{0.23}\\
			\cline{1-11}
			\tiny{native.country} & \tiny{-0.01} & \tiny{0.06} & \tiny{0.00}& \tiny{0.04} &\tiny{0.00}& \tiny{-0.01}& \tiny{0.00}& \tiny{-0.01}& \tiny{1.00}& \tiny{-0.02}\\
			\cline{1-11}
			\tiny{class} & \tiny{0.23} & \tiny{-0.05} & \tiny{0.00}& \tiny{-0.17} &\tiny{-0.21}& \tiny{0.22}& \tiny{0.15}& \tiny{0.23}& \tiny{-0.02}& \tiny{1.00}\\
			\hline
	\end{tabular}}
	\caption{Adult dataset: Pearson Correlation coefficients.}
	\label{table:attCorr}
\end{table}

\begin{table}[!htb]
	\centering
	\setlength{\tabcolsep}{5pt}
	\renewcommand\arraystretch{1.3}
	\resizebox{0.45\textwidth}{5mm}{
	\begin{tabular}{|c|c|c|c|}
		\hline
		Entity & Sensitive attribute& Predicted boundary& Actual boundary\\
		\cline{1-4}
		Sensitive attribute & 1:1  &  -0.20 : -0.16 & -0.21 : -0.21 \\
		\cline{1-4}
		Predicted boundary &  -0.20 : -0.16  & 1:1 & 0.52 : 0.44\\
		\cline{1-4}
		Actual boundary & -0.21 : -0.21 & 0.52 : 0.44 & 1:1 \\
		\cline{1-4}
		\hline
	\end{tabular}}
	\caption{Pearson Correlation coefficients between sensitive attribute, predicted decision boundary and actual decision boundary on Adult dataset. The values before colon are from the HT and after are from FAHT.}
	\label{table: boundaryCorr}
\end{table}

\begin{figure}[H]
	\vspace{-5mm}
	\centering
	\includegraphics[height=0.2\textheight, width=0.4\textwidth]{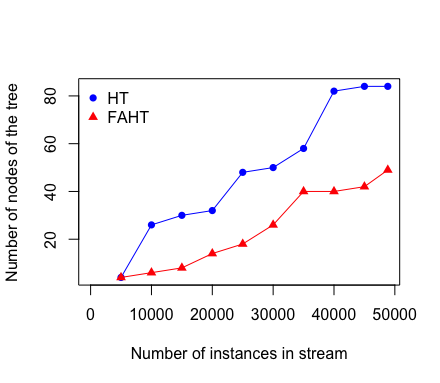}
	\vspace{-1mm}
	\caption{Adult dataset: Model complexity (number of nodes) over the stream.}
	\label{fig:numOfNodes}
\end{figure}

In addition, theoretically, the induction of FAHT is expected to be more conservative than the normal HT. The reason is that the devised fair information gain splitting criterion takes class distribution and discrimination into consideration and is therefore more selective in evaluating candidate splitting attributes. As one can see from Figure~\ref{fig:numOfNodes}, FAHT results in a shorter tree comparing to HT, as its splitting criterion FIG is more restrictive comparing to IG. This conservative characteristic is also helpful for interpretation purpose.


\section{Conclusion}
\label{sec: conclusion}
Although a number of studies have developed approaches to confront with the lack of fairness in data-driven decision-making systems, all of these works tackle fairness as a static problem. We proposed an approach for fairness-aware stream classification, which is able to maintain a moderate  predictive performance with low discrimination scores over the course of the stream. Our approach tackles discrimination by incorporating discrimination-awareness into the model induction process via the fair information gain splitting criterion. Our experiments show that such an integration enables fair predictions while still allowing for model adaptation to the underlying evolving distribution. As part of our future work, we plan to extend our approach for ensemble learning using different window models and learner complexities.
Moreover, we plan to further investigate the similarities and differences between FAHT and HT induced models employing the notion of semantic similarity~\cite{ntoutsi2008general}.  

\bibliographystyle{named}
\bibliography{ijcai19}

\end{document}